\documentclass[conference]{IEEEtran}
\IEEEoverridecommandlockouts
\IEEEpubid{\makebox[\columnwidth]{175-7-8837-8391-8/--/\$--~\copyright~2025 IEEE\hfill} \hspace{\columnsep}\makebox[\columnwidth]{ }}

\usepackage{cite}
\usepackage{amsmath,amssymb,amsfonts}
\usepackage{algorithmic}
\usepackage{graphicx}
\usepackage{textcomp}
\usepackage{xcolor}
 \usepackage{booktabs}
\usepackage{multirow}
\usepackage{graphicx}
\usepackage{subcaption}
\usepackage{hyperref}

\usepackage[utf8]{inputenc}
\usepackage{newunicodechar}
\newunicodechar{−}{\textminus}
\def\BibTeX{{\rm B\kern-.05em{\sc i\kern-.025em b}\kern-.08em
    T\kern-.1667em\lower.7ex\hbox{E}\kern-.125emX}}

\begin{document}

\title{SAM2LoRA: Composite Loss-Guided, Parameter-Efficient Finetuning of SAM2 for Retinal Fundus Segmentation\\
}

    
    
    


\author{\IEEEauthorblockN{1\textsuperscript{st} Sayan Mandal}
\IEEEauthorblockA{
\textit{AMD, San Jose}\\
CA 95124, USA\\
sayanmndl21@gmail.com}
\and
\IEEEauthorblockN{2\textsuperscript{nd} Divyadarshini Karthikeyan}
\IEEEauthorblockA{\textit{Amazon, Toronto} \\
ON M5J0A8, Canada\\
divyadkarthi@gmail.com}
\and
\IEEEauthorblockN{3\textsuperscript{rd} Manas Paldhe}
\IEEEauthorblockA{
\textit{Infinitus Systems, San Francisco}\\
CA 94107, USA\\
manaspaldhe12@gmail.com}
}

\maketitle


\begin{abstract}
We propose SAM2LoRA, a parameter-efficient fine-tuning strategy that adapts the Segment Anything Model 2 (SAM2) for fundus image segmentation. SAM2 employs a masked autoencoder-pretrained Hierarchical Vision Transformer for multi-scale feature decoding, enabling rapid inference in low-resource settings; however, fine-tuning remains challenging. To address this, SAM2LoRA integrates a low-rank adapter into both the image encoder and mask decoder, requiring fewer than 5\% of the original trainable parameters. Our analysis indicates that for cross-dataset fundus segmentation tasks, a composite loss function combining segmentationBCE, SoftDice, and FocalTversky losses is essential for optimal network tuning. Evaluated on 11 challenging fundus segmentation datasets, SAM2LoRA demonstrates high performance in both blood vessel and optic disc segmentation under cross-dataset training conditions. It achieves Dice scores of up to 0.86 and 0.93 for blood vessel and optic disc segmentation, respectively, and AUC values of up to 0.98 and 0.99, achieving state-of-the-art performance while substantially reducing training overhead. \footnote{This research was conducted independently of the authors' affiliations and reflects solely the views and efforts of the authors.}\footnote{The source code is available at \href{https://github.com/sayanmndl/SAM2LoRA}{\texttt{sayanmndl/SAM2LoRA}}.}

\end{abstract}


\section{Introduction}
Fundus imaging is a noninvasive examination in which images of the fundus (inner back wall of the eye, including the retina, macula, optic nerves, and choroid) are captured to diagnose and treat various ocular diseases such as Macular Degeneration (AMD) \cite{Fleckenstein2021AgeRelatedMacularDegeneration} diabetic retinopathy (DR) \cite{Antonetti2021MolecularCellularDiabeticRetinopathy}, and glaucoma \cite{Casson2012DefinitionGlaucoma, mandal2021assessing}. Diagnosis primarily relies on analyzing the retinal blood vessels and optic discs, two critical regions for assessing ocular conditions. Manual analysis of fundus image is cumbersome, labor intensive, time consuming, and error prone, making computational analysis, especially ocular segmentation, essential for both diagnosis and the development of advanced applications \cite{besenczi2016review}.

Image segmentation is a fundamental task in computer vision. With the advent of deep learning, segmentation models have increasingly utilized deep neural networks \cite{huang2022fully}. Numerous methods have been developed for ocular segmentation that effectively capture features relevant to both blood vessel and optic disc segmentation. The introduction of models based on convolutional neural networks (CNN) eliminated the need of manual feature extraction, benefited from pre-trained models (transfer learning), and outperformed previous approaches \cite{septiarini2023automatic}. U-Net, a seminal model for medical segmentation, further popularized deep learning approaches and inspired subsequent modifications that improved performance \cite{huang2022fully}. More recently, attention-based models, particularly transformer architectures, have enhanced segmentation performance by integrating attention mechanisms, enabling the development of foundational models pretrained on extensive datasets that can be adapted to domain-specific tasks with minimal adjustments \cite{xiao2023transformers}. 

Recently, new foundational image segmentation models have been introduced and adopted in various fields. Meta’s Segment Anything (SAM) and Segment Anything 2 (SAM2) models have gained mainstream attention due to their performance, flexibility, and ease of use \cite{ravi2024sam2segmentimages}. It has also become a popular model for medical image segmentation. SAM2 variants have been used to segment tumors, organs, lesions, and other structures \cite{zhang2024unleashingpotentialsam2biomedical}, as well as segments of the optic disc and retinal vessels \cite{zhu2024medicalsam2segment}. Although most applications focus on adjusting models for general medical image segmentation, fine-tuning SAM2 for domain-specific tasks such as fundus image segmentation with intricate blood vessel and optic disc details remains under-explored.

SAM2 has millions of parameters, and hence fine-tuning it is computationally expensive. The low-rank adaptation technique (LoRA) is frequently used \cite{hu2021lora4} to efficiently fine-tune such models. LoRA assumes that the change in weights ($\Delta W$) during fine-tuning is of low rank. Consequently, only a small fraction of the weights need to be updated, resulting in a faster and more efficient fine-tuning process. In this paper, we leverage LoRA to fine-tune SAM2 for fundus image segmentation and demonstrate that this strategy improves state-of-the-art performance. Our contributions are as follows.

\begin{itemize}
    \item We propose SAM2LoRA, a fine-tuning strategy that adapts SAM2 for fundus image segmentation by applying LoRA to both the image encoder and mask decoder for domain specialization.
    \item We introduce composite loss-guided tuning, which combines segmentationBCE, SoftDice, and FocalTversky losses to optimize the network for complex segmentation tasks, such as blood vessel segmentation, without compromising performance on optic disc segmentation.
    \item We present experimental results showing that the proposed strategy reaches current state-of-the-art methods in terms of Dice and AUC scores.
\end{itemize}

\section{Methods}

\subsection{SAM2 Model Overview}

The SAM2 is an enhancement over SAM, designed for efficient and accurate image segmentation. It is built on a hierarchical vision transformer (Hiera) architecture that leverages masked autoencoders (MAE) for pre-training, enabling robust representation learning from large unlabeled datasets. A core feature of SAM2 is its multi-scale feature extraction, which is crucial for segmenting structures of varying sizes, such as blood vessels and the optic disc in fundus images \cite{chen2024hiera, kirillov2023segment}. The SAM2 architecture comprises three primary components:
\begin{itemize}
    \item \textbf{Image Encoding:} The image encoder extracts feature maps at multiple scales using the Hiera vision transformer. This hierarchical model processes input images at resolutions of 1/4, 1/8, 1/16, and 1/32 with adaptive attention windows, thereby capturing both global context and fine-grained details in a single forward pass. This multi-scale approach is key to achieving high accuracy in complex segmentation tasks, such as those encountered in ophthalmology \cite{ravi2024sam2segmentimages}.
    \item \textbf{Mask Decoding:} The mask decoder transforms the encoder’s output into segmentation masks using multiple attention layers to refine the segmentation. This process ensures the precise delineation of structures like blood vessels and optic discs in fundus images \cite{kirillov2023segment}.
    \item \textbf{Prompt Encoding:} Similar to SAM, SAM2 employs a series of transposed convolution and convolution layers to encode input prompts such as image masks, points, or bounding boxes thereby focusing the segmentation on specific regions of the image.
\end{itemize}


\subsection{Low Ranked Adaptation}

\begin{figure*}[!t]  
\centering  
\includegraphics[width=0.9\textwidth]{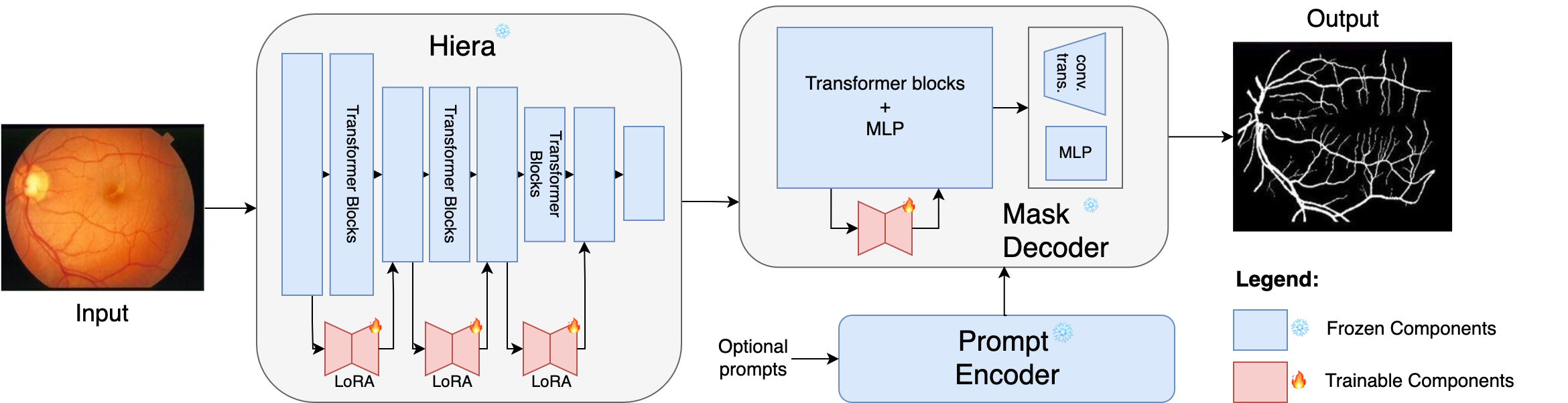}  
\vspace{-0.2cm}  
\caption{SAM2 LoRA Framework Architecture: LoRA is integrated into every projection layer within the transformer blocks' attention modules.}
\label{fig:sam2_arch}
\vspace{-0.3cm}  
\end{figure*}

Low-Rank Adaptation or LoRA reduces the computational burden of fine-tuning large models by introducing low-rank matrices into the model’s weight matrices. LoRA modules are integrated into the vision transformer architecture by modifying the weight matrices of the linear projections within the attention modules of both the encoder and the decoder (Figure \ref{fig:sam2_arch}). Specifically, LoRA introduces low-rank approximations of the full weight adapters in these modules, enabling efficient fine-tuning. This significantly reduces the number of trainable parameters compared to fine-tuning the entire model. 

\subsection{Composite Loss}

Fundus image segmentation presents challenges at both the micro level (blood vessel segmentation) and the macro level (optic disc segmentation). To address these challenges, SAM2 employs a composite loss function that combines three loss components: segmentation binary cross-entropy (segmentationBCE), SoftDice loss, and FocalTversky loss. Each loss function serves a distinct purpose:
\begin{itemize}
    \item \textbf{SegmentationBCE:} The binary cross-entropy loss performs pixel-wise classification, enabling the model to assign each pixel to one of the predefined classes. This loss is particularly effective for dense, pixel-level segmentation tasks \cite{long2015fully7}.
    \item  \textbf{SoftDice Loss:} Dice loss directly optimizes the overlap between predicted and ground truth masks. It encourages the accurate segmentation of smaller structures, such as blood vessels or the optic cup, by emphasizing their precise delineation. A small constant ($\epsilon$), typically set to $1e^{-5}$, is added to prevent division by zero \cite{milletari2016v}.
    \item \textbf{FocalTversky Loss:} FocalTversky loss addresses class imbalance by assigning greater weight to hard-to-segment regions, thus improving the segmentation of underrepresented structures like small blood vessels \cite{abraham2019novel}.
\end{itemize}

\section{Experiments}

\subsection{Datasets}

We employ a collection of datasets for our segmentation tasks. For blood vessel segmentation, we use five datasets: CHASEDB1\cite{fraz2012ensemble}, DRIVE\cite{staal2004ridge}, FIVES\cite{jin2022fives}, HRF\cite{budai2013robust}, and STARE\cite{hoover2000locating}. For optic disc segmentation, we use six datasets: DRISHTIGS\cite{6867807}, G1020\cite{bajwa2006g1020}, GRAPE\cite{huang2023grape}, ORIGA\cite{zhang2010origa}, PAPILADB\cite{kovalyk2022papila}, and REFUGE2\cite{fang2202refuge2}. Each dataset consists of high-quality fundus images paired with binary masks delineating the relevant structures (e.g., blood vessels, optic disc, and optic cup). For our experiments, we focus on blood vessel and optic disc segmentation. All datasets were curated by removing duplicate samples and discarding cases with missing segmentation masks. When available, we used the provided training–test splits by the author; otherwise, we applied a random split (25\% test) with a fixed seed. The training subsets for each task were combined for cross-dataset fine-tuning of the corresponding segmentation models, while evaluation was conducted separately on each dataset. In total, there are 687 training samples and 246 testing samples for blood vessel segmentation, and 2541 training samples and 1149 testing samples for optic disc segmentation.

\subsection{Implementation Details}

To augment the limited data and increase variability, we applied standard transformation techniques commonly used in U-Net training. These augmentations include horizontal and vertical flips, random perspective transformations, resized cropping, rotations, and affine transformations. Additionally, color augmentations, such as color jitter, Gaussian blur, and sharpness adjustments were applied sparingly to enhance model robustness. All fundus images were upsampled to $1000 \times 1000$ pixels and normalized, with corresponding binary segmentation masks generated to ensure uniformity during training. Although SAM2 inherently predicts three segmentation masks, we fine-tuned only the mask corresponding to the task at hand due to the lack of datasets with simultaneous annotations for both blood vessels and optic discs.

LoRA was integrated into the "$hiera\_l$" architecture for both the image encoder and mask decoder by exploring ranks of 8, 16, 32 and 64. LoRA was applied uniformly to all query, key, value, and output projection layers, while other components, such as the prompt encoder, remained frozen. Although SAM2 contains an additional memory head, it was excluded from training and evaluation because our focus is solely on image segmentation. In all cases, the LoRA alpha parameter was set to twice the rank, according to established recommendations\cite{lu2023empirical}.

Our training loss is an equal weighted linear combination of SegmentationBCE, Soft Dice loss, and Focal Tversky loss. This composite loss was selected after extensive empirical analysis to address the severe class imbalance typical in medical segmentation masks (where segmented regions often cover less than 5\% of the image area). Specifically, Soft Dice loss optimizes structural overlap via the Dice coefficient outperforming binary cross-entropy (BCE) \cite{milletari2016v}, while Focal Tversky loss leverages the Tversky index with focal scaling to balance false positives and false negatives, thereby stabilizing training and enhancing performance \cite{abraham2019novel}. Optimization was performed using the AdamW optimizer with a learning rate of $1 \times 10^{-4}$, weight decay of $3 \times 10^{-4}$, and momentum parameters $\beta_1 = 0.9$ and $\beta_2 = 0.999$. A cosine annealing schedule with warm restarts (warm-up step size = 1000) was employed, and training was carried out for 20,000 steps with a batch size of 3 on a single Nvidia RTX 4070 Ti GPU until convergence. Given that SAM2 is a prompt-based segmentation model, we fine-tuned the LoRA parameters under randomized prompt scenarios to fully characterize its performance.

\subsection{Quantitative Analysis}

\begin{table*}[!t]
  \centering
  \caption{Performance of \textbf{SAM2LoRA} versus leading vessel-segmentation methods (Dice / AUC ↑). “--” means result not reported.}
  \label{tab:vessel_res}
  \setlength{\tabcolsep}{5pt}  

  \begin{tabular}{lcccccccccc}
    \toprule
    \multirow{2}{*}{\textbf{Model}} &
      \multicolumn{2}{c}{\textbf{DRIVE}}   &
      \multicolumn{2}{c}{\textbf{STARE}}   &
      \multicolumn{2}{c}{\textbf{CHASEDB1}}&
      \multicolumn{2}{c}{\textbf{HRF}}     &
      \multicolumn{2}{c}{\textbf{FIVES}}   \\
    \cmidrule(lr){2-3}\cmidrule(lr){4-5}\cmidrule(lr){6-7}\cmidrule(lr){8-9}\cmidrule(lr){10-11}
     & Dice & AUC & Dice & AUC & Dice & AUC & Dice & AUC & Dice & AUC \\
    \midrule
    SE U-Net \cite{shen2022self}          & --   & --   & --   & --   & --   & --   & \textbf{0.84} & \textbf{0.99} & --   & --   \\
    MFA U-Net \cite{cao2023mfa}           & \textbf{0.83} & 0.98 & \textbf{0.84} & \textbf{0.99} & \textbf{0.84} & \textbf{0.99} & 0.81 & \textbf{0.99} & 0.84 & 0.98 \\
    FS Rep. Net \cite{seo2025full}        & \textbf{0.83} & \textbf{0.99} & --   & --   & --   & --   & --   & --   & --   & --   \\
    \textbf{SAM2LoRA (ours)}              & 0.82 & 0.98 & 0.81 & 0.97 & 0.83 & \textbf{0.99} & 0.78 & 0.98 & \textbf{0.87} & \textbf{0.99} \\
    \bottomrule
  \end{tabular}
\end{table*}


\begin{table*}[!t]
  \centering
  \caption{Comparison of \textbf{SAM2LoRA} with leading optic-disc-segmentation models (Dice score ↑). “--” denotes results not reported.  No public Dice comparison is available for GRAPE except ours.}
  \label{tab:od_res}
  \setlength{\tabcolsep}{6pt}  

  \begin{tabular}{lcccccc}
    \toprule
    \multirow{2}{*}{\textbf{Model}} &
      \multicolumn{6}{c}{\textbf{Dataset}} \\ 
    \cmidrule(lr){2-7}
     & \textbf{DRISHTI} & \textbf{REFUGE} & \textbf{G1020} & \textbf{GRAPE} & \textbf{ORIGA} & \textbf{PAPILA} \\
    \midrule
    SAM GT \cite{yii2023data}               & 0.96 & --   & --   & --   & --   & 0.95 \\
    R-Bend \cite{10551123}                  & --   & --   & --   & --   & \textbf{0.97} & --   \\
    Attn U-Net \cite{chen2024optic}         & 0.96 & --   & --   & --   & --   & --   \\
    M-Net \cite{tadisetty2023identifying}   & --   & \textbf{0.97} & --   & --   & --   & --   \\
    Mask R-CNN \cite{bajwa2020g1020}        & --   & --   & 0.86 & --   & --   & --   \\
    MedSAM2 \cite{sam_zhu2024medical}   & --   & 0.80 & --   & --   & --   & --   \\
    \textbf{SAM2LoRA (ours)}                & \textbf{0.97} & 0.96 & \textbf{0.94} & \textbf{0.96} & 0.95 & \textbf{0.96} \\
    \bottomrule
  \end{tabular}
\end{table*}

In Table I and Table II we have reported the Dice coefficient and AUC scores for all datasets, which reflect the optimal performance achieved across LoRA ranks. Our results are benchmarked against state-of-the-art (SOTA) methods for each dataset. It is important to note that SAM2LORA performance is computed without fine tuning for the specific dataset, whereas the other models were fine tuned for the datasets. In spite of this handicap, SAM2LoRA consistently meets or exceeds the performance of contemporary approaches. For blood vessel segmentation, SAM2LoRA demonstrates near-SOTA performance compared to methods such as MFA U-Net\cite{cao2023mfa} and Full Scale Representation Net\cite{seo2025full} for the DRIVE, STARE, and CHASEDB1 datasets, and it outperforms MFA U-Net on the FIVES dataset (Table \ref{tab:vessel_res}). However, we note that SAM2LoRA's performance on the HRF dataset was lower, as it did not achieve the Dice and AUC scores of Squeeze Excitation U-Nets\cite{shen2022self}. For optic disc segmentation, SAM2LoRA exhibits competitive performance across multiple datasets (Table \ref{tab:od_res}). SAM2LoRA achieves SOTA performance on the DRISHTI-GS, G1020, and PAPILA datasets, outperforming SAM GT and MedSAM2\cite{yii2023data,sam_zhu2024medical} (SAM fine tuned using ground truth data), Attention U-Net\cite{chen2024optic} (enhanced U-Net architecture with attention and residual connections), and Mask R-CNN\cite{bajwa2020g1020} (a two-stage instance segmentation model adapted to detect and segment the optic disc and cup in fundus images) methods, respectively. It also demonstrates performance comparable to that of the M-Net\cite{tadisetty2023identifying} and R-Bend\cite{10551123} networks, which represent the current SOTA for the REFUGE and ORIGA datasets. Furthermore, for the GRAPE dataset, where no prior optic disc segmentation model exists, our method achieves a Dice score of 0.96, reaching SOTA. This consistent performance across diverse datasets, despite the model being trained in a cross-dataset fashion, demonstrates the robustness and generalizability of SAM2LoRA for both blood vessel and optic disc segmentation tasks.

\subsection{Qualitative Analysis}

\begin{figure*}[!t]
    \centering
    
    \begin{minipage}[b]{0.95\textwidth}
        \centering
        \includegraphics[width=0.95\textwidth]{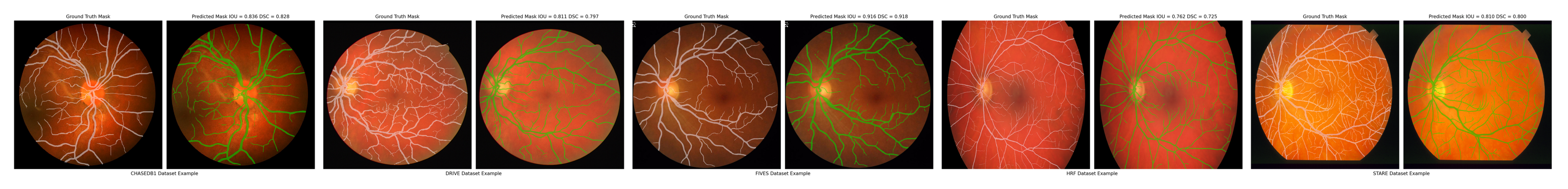}
        \vspace{0.1cm}
        \centerline{\fontsize{8}{10}\selectfont (a) Qualitative blood-vessel segmentation results on five retinal datasets.}
    \end{minipage}
    
    \vspace{0.3cm}
    
    \begin{minipage}[b]{0.95\textwidth}
        \centering
        \includegraphics[width=0.95\textwidth]{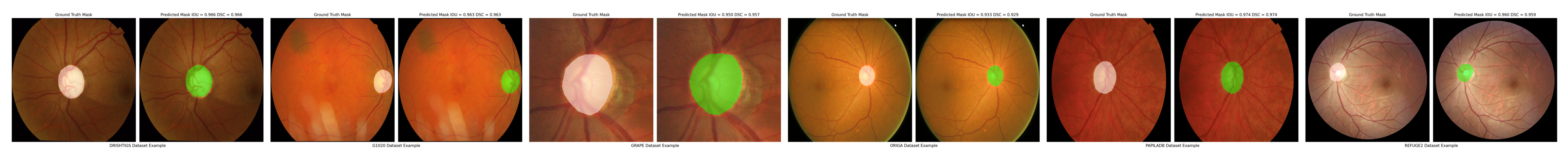}
        \vspace{0.1cm}
        \centerline{\fontsize{8}{10}\selectfont (b) Qualitative optic disc segmentation results on six retinal datasets.}
    \end{minipage}
    
    \vspace{0.2cm}
    \caption{Side-by-side comparison of segmentation masks: ground truth versus predictions from SAM2LoRA at LoRA rank 32.}
    \label{fig:qualitative}
\end{figure*}

Figure \ref{fig:qualitative} presents representative segmentation maps for both blood vessel and optic disc tasks across all datasets. In these visualizations, fundus images are overlaid with ground truth and predicted masks in distinct colors. The results indicate that optic disc boundaries are clearly delineated, with predictions closely matching the ground truth. Although blood vessel segmentation is particularly challenging, the model demonstrates robust performance in regions of thicker vasculature, with some degradation in finer branches. Overall, the qualitative outcomes are promising, especially given that a single model was applied uniformly across diverse datasets.

Notably, SAM2LoRA produces high-quality blood vessel segmentations without prompting, whereas optic disc segmentation may require prompting. For datasets with smaller optic discs, a single positive point prompt yields high-quality maps, while for cases with larger optic discs (e.g., the GRAPE dataset), a box prompt is necessary. These observations suggest potential limitations of SAM2LoRA in handling varying optic disc sizes, although high-quality segmentation masks are ultimately achieved. While the architecture is unchanged, we show that LoRA applied simultaneously to the image encoder and mask decoder, trained with uniformly weighted Soft-Dice, binary cross-entropy, and focal-Tversky losses, obtains SOTA performance on retinal segmentation task while modifying $< 5\%$ of the network’s parameters. This configuration (LoRA rank 16–32) constitutes a reproducible, compute and memory efficient alternative to full retraining.

\begin{figure*}[t]
  \centering
  \begin{subfigure}[b]{0.49\textwidth}
    \centering
    \includegraphics[width=\linewidth]{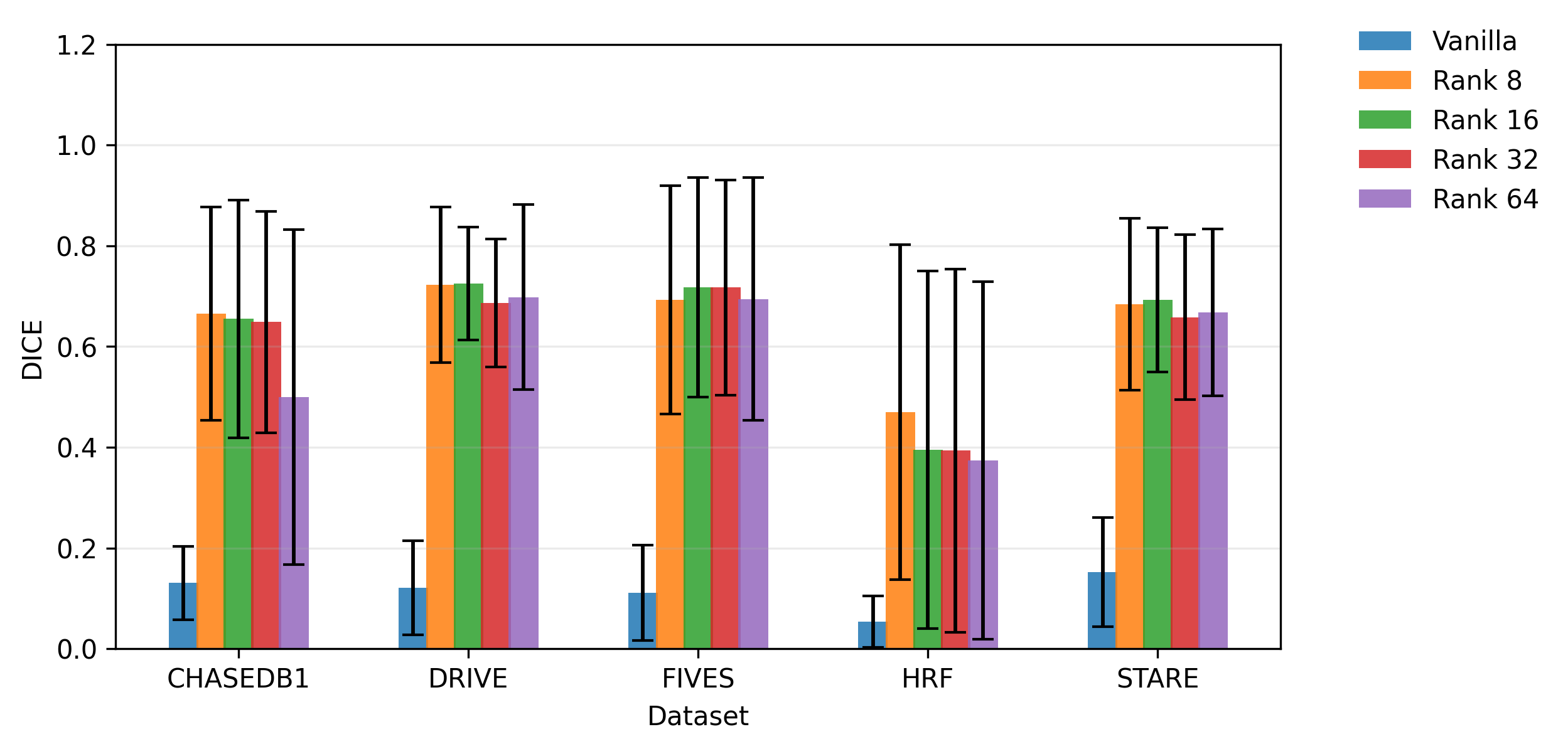}
    \caption{Blood-vessel segmentation examples}
    \label{fig:rank_vessel}
  \end{subfigure}
  \hfill
  \begin{subfigure}[b]{0.49\textwidth}
    \centering
    \includegraphics[width=\linewidth]{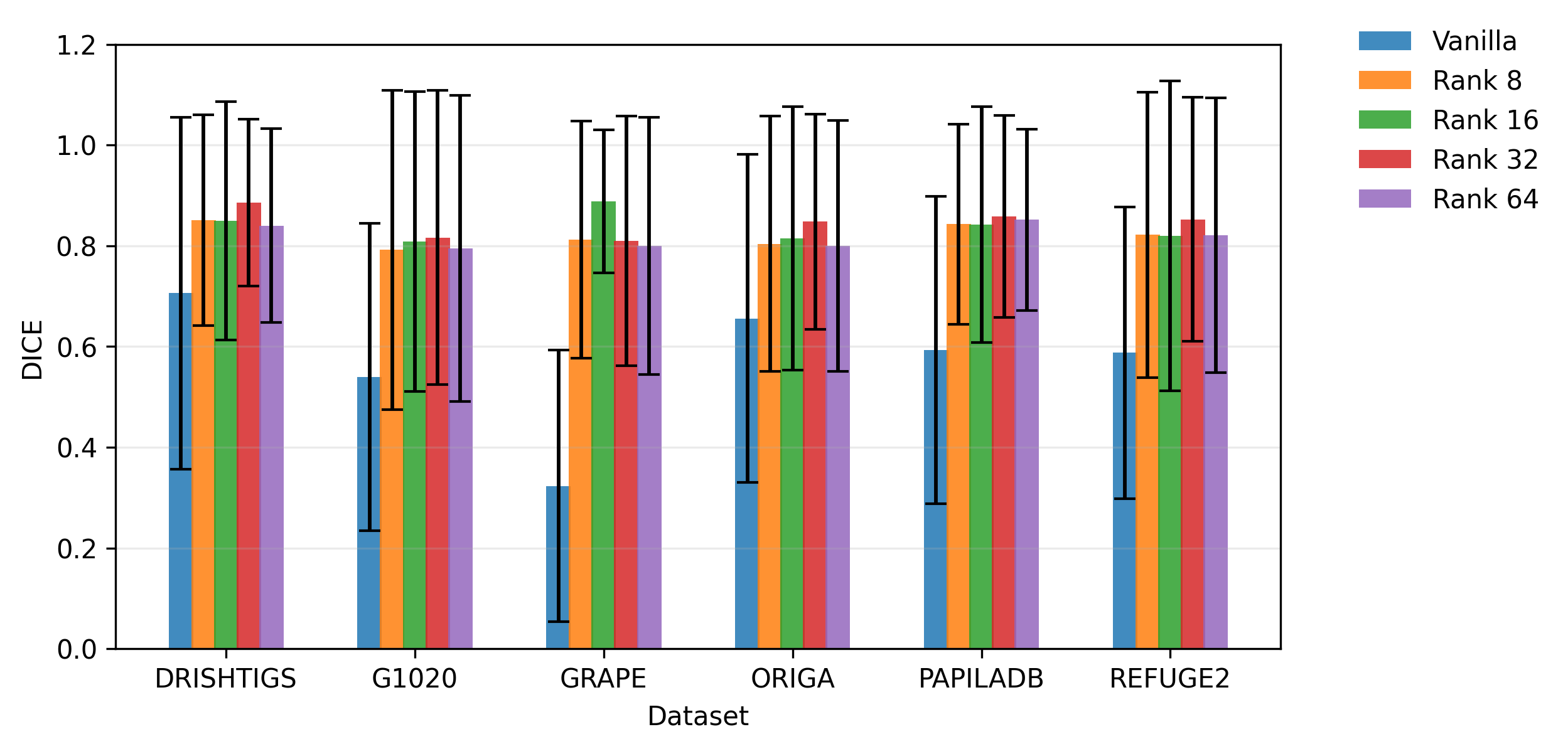}
    \caption{Optic-disc segmentation examples}
    \label{fig:rank_disc}
  \end{subfigure}

  \caption{Dice scores for different LoRA ranks compared across datasets aggregated over all prompt modes.}
  \label{fig:dice_auc_main}
\end{figure*}

\begin{table}[h]
  \centering
  \caption{Segmentation performance (Dice score) across LoRA ranks (8, 16, 32, 64) and six ablation modes as described in Section \ref{sec:ablation}.}
  \label{tab:ablation}

  \setlength{\tabcolsep}{4pt}
  \footnotesize
  \renewcommand{\arraystretch}{1.1}

  \begin{subtable}{\linewidth}
    \centering
    \caption{Vessel segmentation}
    \label{tab:vessel_seg}
    \resizebox{\linewidth}{!}{%
      \begin{tabular}{lcccc}
        \toprule
        \textbf{Rank $\rightarrow$} & \multirow{2}{*}{\textbf{8}} & \multirow{2}{*}{\textbf{16}}
                      & \multirow{2}{*}{\textbf{32}} & \multirow{2}{*}{\textbf{64}} \\
        \textbf{Ablation $\downarrow$} &&&& \\ \midrule
        0 & $0.682 \pm 0.233$ & $0.699 \pm 0.232$ & $0.694 \pm 0.229$ & $0.671 \pm 0.258$ \\
        1 & $0.619 \pm 0.256$ & $0.663 \pm 0.248$ & $0.682 \pm 0.239$ & $0.697 \pm 0.230$ \\
        2 & $0.633 \pm 0.245$ & $0.648 \pm 0.262$ & $0.627 \pm 0.265$ & $0.745 \pm 0.229$ \\
        3 & $0.672 \pm 0.244$ & $0.672 \pm 0.242$ & $\mathbf{0.757 \pm 0.211}$ & $\mathbf{0.750 \pm 0.253}$ \\
        4 & $\mathbf{0.707 \pm 0.207}$ & $\mathbf{0.735 \pm 0.199}$ & $0.658 \pm 0.225$ & $0.590 \pm 0.264$ \\
        5 & $0.211 \pm 0.194$ & $0.243 \pm 0.221$ & $0.256 \pm 0.215$ & $0.335 \pm 0.250$ \\
        \bottomrule
      \end{tabular}
    }
  \end{subtable}
  
  \vspace{0.8em} 
  
  \begin{subtable}{\linewidth}
    \centering
    \caption{Optic disc segmentation}
    \label{tab:optic_disc}
    \resizebox{\linewidth}{!}{%
      \begin{tabular}{lcccc}
        \toprule
        \textbf{Rank $\rightarrow$} & \multirow{2}{*}{\textbf{8}} & \multirow{2}{*}{\textbf{16}}
                      & \multirow{2}{*}{\textbf{32}} & \multirow{2}{*}{\textbf{64}} \\
        \textbf{Ablation $\downarrow$} &&&& \\ \midrule
        0 & $0.815 \pm 0.271$ & $\mathbf{0.829 \pm 0.272}$ & $\mathbf{0.840 \pm 0.245}$ & $\mathbf{0.813 \pm 0.264}$ \\
        1 & $0.782 \pm 0.283$ & $0.805 \pm 0.280$ & $0.798 \pm 0.282$ & $0.785 \pm 0.263$ \\
        2 & $\mathbf{0.820 \pm 0.285}$ & $0.822 \pm 0.259$ & $0.818 \pm 0.267$ & $0.790 \pm 0.290$ \\
        3 & $0.815 \pm 0.283$ & $0.807 \pm 0.286$ & $0.793 \pm 0.301$ & $0.774 \pm 0.298$ \\
        4 & $0.789 \pm 0.278$ & $0.798 \pm 0.268$ & $0.796 \pm 0.264$ & $0.796 \pm 0.248$ \\
        5 & $0.701 \pm 0.306$ & $0.699 \pm 0.304$ & $0.710 \pm 0.295$ & $0.713 \pm 0.296$ \\
        \bottomrule
      \end{tabular}
    }
  \end{subtable}

\end{table}
\vspace{-0.8em}

\begin{figure}[ht]
  \centering
  
  \begin{subfigure}{0.49\textwidth}
    \includegraphics[width=\linewidth]{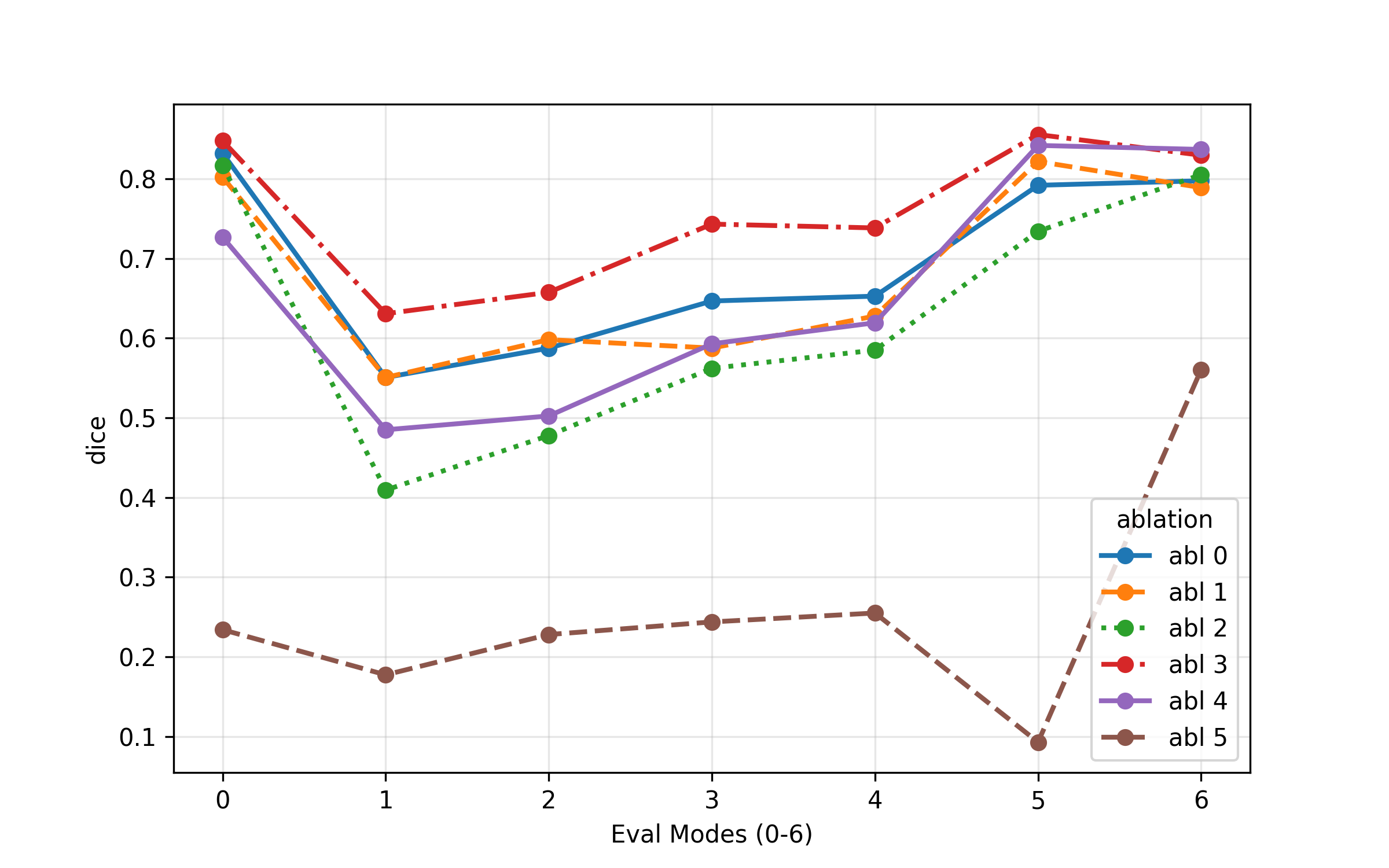} 
    \caption{Rank 32}
    \label{fig:sub1}
  \end{subfigure}
  \hfill
  \begin{subfigure}{0.49\textwidth}
    \includegraphics[width=\linewidth]{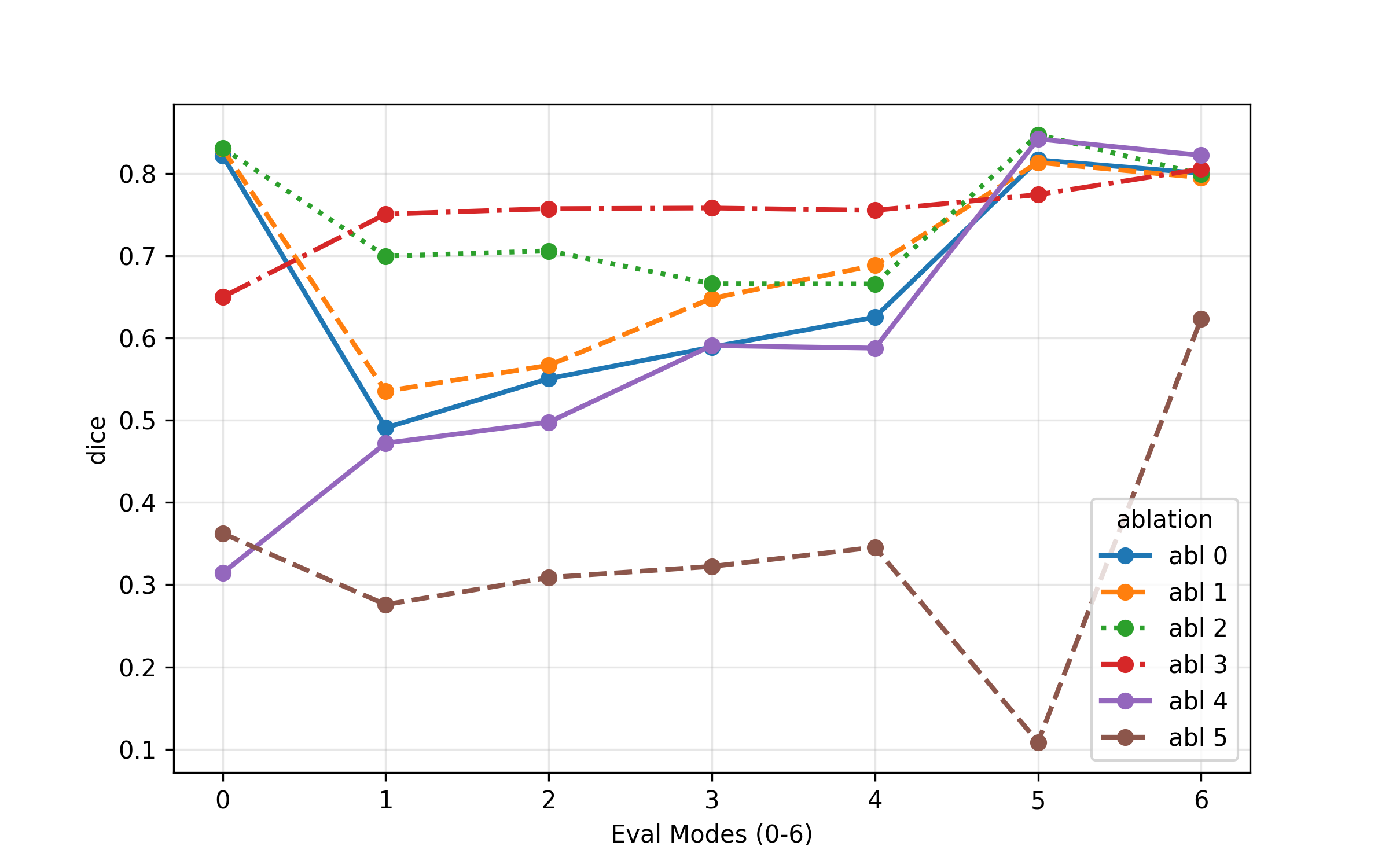}
    \caption{Rank 64}
    \label{fig:sub2}
  \end{subfigure}
  
  \caption{Vessel Segmentation Dice performance under varying prompt scenarios, reported at each ablation mode’s best-performing LoRA rank, with ablation modes defined as in Section \ref{sec:ablation}}
  \label{fig:vs_ablation}
\end{figure}

\begin{figure}[ht]
  \centering
  
  \begin{subfigure}{0.49\textwidth}
    \includegraphics[width=\linewidth]{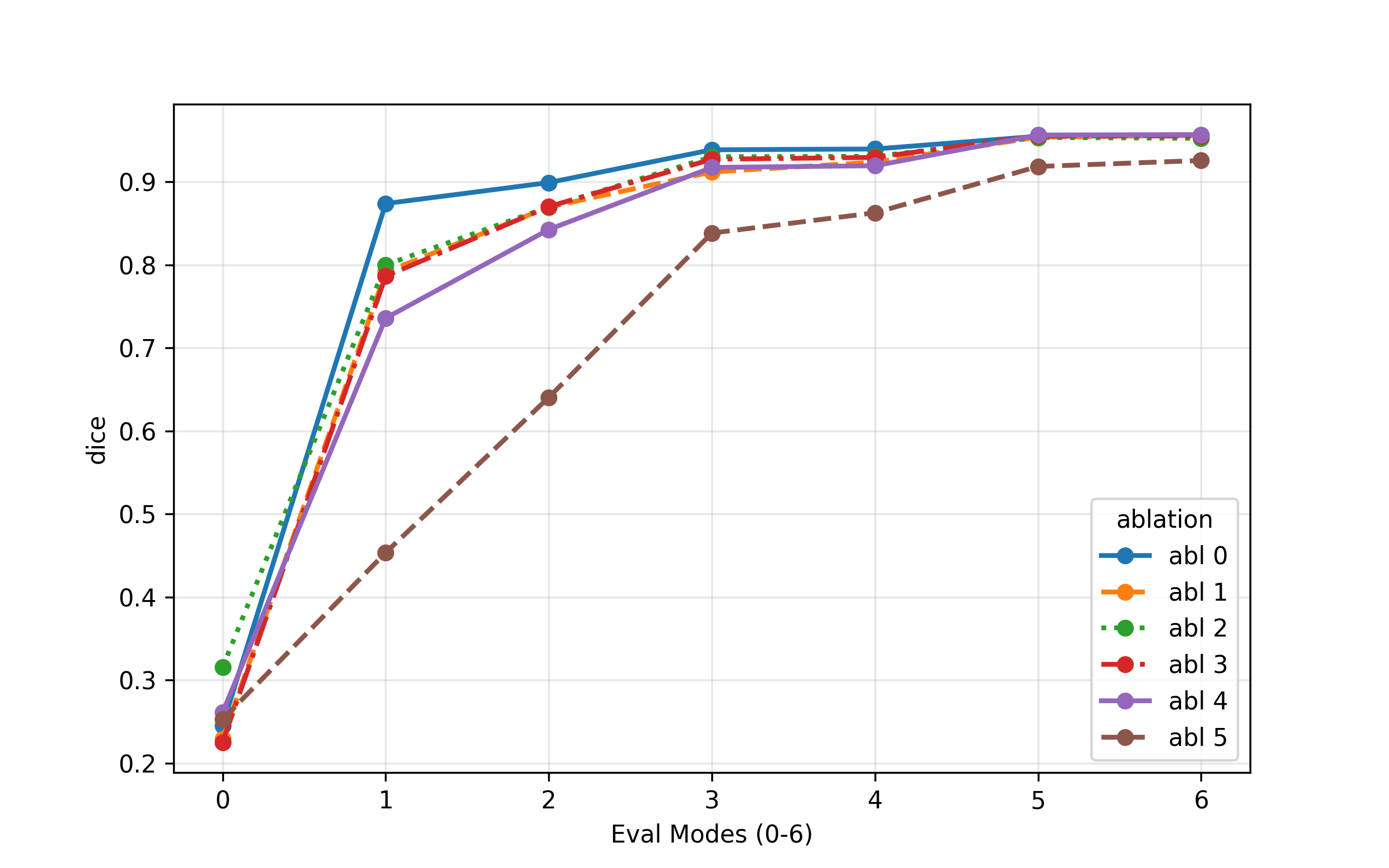} 
    \caption{Rank 16}
    \label{fig:sub1}
  \end{subfigure}
  \hfill
  \begin{subfigure}{0.49\textwidth}
    \includegraphics[width=\linewidth]{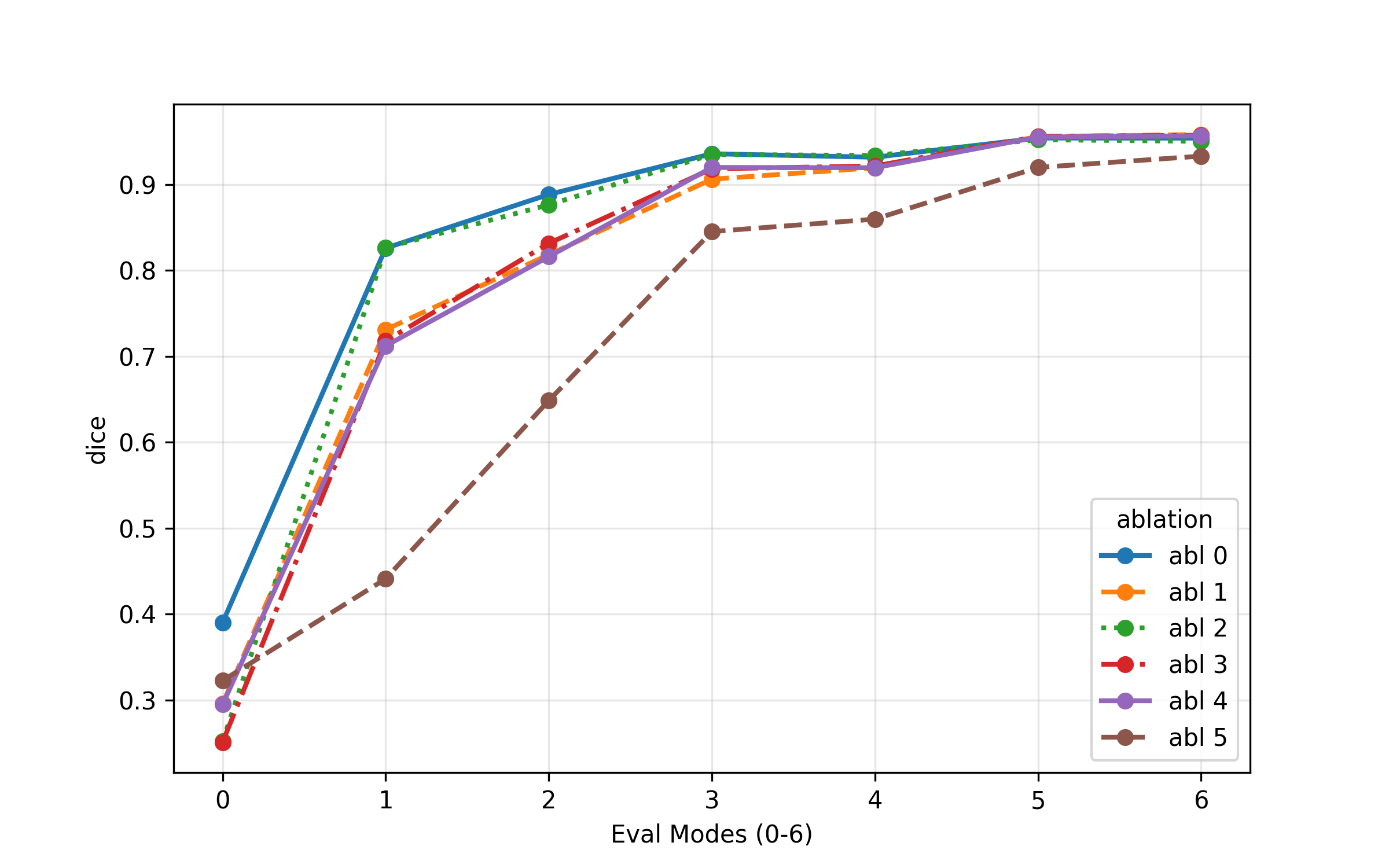}
    \caption{Rank 32}
    \label{fig:sub2}
  \end{subfigure}
  
  \caption{Optic Disc Segmentation Dice performance under varying prompt scenarios, reported at each ablation mode’s best-performing LoRA rank, with ablation modes defined as in Section \ref{sec:ablation}}
  \label{fig:od_ablation}
\end{figure}

\section{Ablation Study}
\label{sec:ablation}

We compared the \emph{vanilla} SAM model with LoRA variants, sweeping four LoRA ranks (8, 16, 32, and 64) across six ablation configurations: (abl-0) the composite loss SoftDice, BCE, and focal-Tversky with LoRA applied to \emph{both} the image encoder and the mask decoder; (abl-1) BCE-only loss; (abl-2) focal-Tversky-only loss; (abl-3) Soft-Dice-only loss; (abl-4) LoRA on the encoder only; and (abl-5) LoRA on the decoder only. Prompt sensitivity was studied under seven scenarios: (eval-0) no prompts; (eval-1) a single positive random point; (eval-2) two positive random points; (eval-3) five positive random points; (eval-4) five positive and one negative random point; (eval-5) a single box prompt; and (eval-6) a combination of five positive points with one box prompt. All other hyper-parameters were held constant to ensure fair comparison.

\subsection{LoRA Ranks, Module Placement, and Loss-Functions}

The results (Figure \ref{fig:dice_auc_main}) indicate that a LoRA rank of approximately 16-32 range provides the best trade-off between performance and parameter efficiency, as higher ranks yield marginal improvements at the cost of increased parameters. Ablation studies showed that the complete setting SoftDice + BCE + Focal Tversky loss with LoRA applied to both the image encoder and the mask decoder delivers the most reliable performance (Table \ref{tab:ablation}. Paired two sided (t)-tests across ranks indicated that removing the BCE term reduced the mean Dice for optic-disc segmentation by 3.2 points $(95 \% CI [−4.4, −1.9], (p = 0.004))$, whereas dropping Soft-Dice or Focal Tversky produced no significant change $(p > 0.05)$. No loss-only ablation significantly affected the retinal-vessel task $(p > 0.34)$. Eliminating LoRA from the image encoder caused the sharpest degradation, lowering Dice by 11.9 points for optic discs $(95 \% CI [−14.2, −9.6], (p < 0.001))$ and by 42.5 points for vessels $(95 \% CI [−52.2, −32.8], (p < 0.001))$.

When disc and vessel results were combined (eight paired observations), the complete configuration maintained a 2.6-point average advantage over the best reduced variant $(p = 0.023)$. The minimum Dice achieved with the full recipe remained above 0.813 for optic discs and 0.671 for vessels, whereas the decoder-only LoRA variant fell to 0.699 and 0.211, respectively. These findings indicate that each loss component and each LoRA branch mitigates distinct failure modes; removing any single element yields a statistically significant drop on at least one task. Consequently, the full configuration is the most dependable choice for simultaneous optic-disc and retinal-vessel segmentation.

\subsection{Sweeping Prompt Modes}

Analysis of the prompt configurations reveals that prompting benefits well-defined structures such as the optic disc. A single positive point prompt is optimal for smaller optic discs, while a box prompt is preferable for larger optic discs. Negative point prompts did not significantly affect segmentation masks. For blood vessel segmentation, the absence of prompts produced superior results, likely due to the complex and diffuse nature of vascular structures. In some cases, the introduction of point prompts degraded performance, suggesting that the prompt encoder might require further fine-tuning for blood vessel segmentation tasks.

Additionally, the box prompts did not offer significant performance gains. The prompt sensitivity curves for optic-disc segmentation (ranks 16 and 32) confirm that the complete configuration (abl-0) maintains the highest Dice across evaluation modes, with the advantage being most pronounced when only a small number of prompts is supplied (modes 0–2) (Figure \ref{fig:od_ablation}). As the prompt budget increases, the gap narrows and, by mode 6, all ablations except the decoder-only LoRA variant (abl-5) converge above 0.92. The encoder-only LoRA setting (abl-4) initially lags but catches up once three or more prompts are provided, suggesting that decoder adaptation is vital when contextual information is limited. Overall, the full recipe yields a consistently steeper improvement curve and the highest plateau value, indicating superior robustness to prompt variability.

Retinal-vessel segmentation displays a different pattern. Across ranks 32 and 64, most configurations experience an early decline in Dice at mode 1, highlighting the sensitivity of thin vascular structures to sub-optimal prompts (Figure \ref{fig:vs_ablation}. The Soft-Dice-only loss (abl-3) partially offsets this dip and leads the field through modes 0–5, yet the complete configuration closes the gap at mode six and outperforms abl-3 at the highest rank. Decoder-only LoRA (abl-5) persistently underperforms, never exceeding 0.35 before the maximum prompt budget and still trailing the other settings by at least fifteen percentage points at mode 6. Taken together, these curves reinforce the table-based findings: the combined loss with dual-branch LoRA provides the most stable response to varying prompt regimes while safeguarding worst-case performance for both compact discs and complex vessel trees.

\section{Conclusion}

In this work, we introduced SAM2LoRA, a LoRA adaptation of SAM2 for fundus image segmentation. By leveraging a composite loss comprising segmentationBCE, SoftDice, and FocalTversky loss, we fine-tuned SAM2 on a cross-dataset training set spanning multiple datasets for blood vessel and optic disc segmentation. We integrated LoRA into the attention modules of both the image encoder and mask decoder, demonstrating that SAM2LoRA achieves SOTA performance in both tasks while maintaining a low trainable parameter footprint (less than 5\% of parameters). Comprehensive quantitative and qualitative evaluations show that SAM2LoRA outperforms established methods and exhibits robust segmentation across diverse prompting scenarios. Ablation studies further confirm the optimal configuration of LoRA rank and prompt strategies, demonstrating the value of LoRA for efficient domain adaptation in foundational segmentation models. These findings have significant implications for clinical fundus imaging, enabling more efficient and accurate diagnostic workflows.

\bibliography{ref}

\begin{thebibliography}{10}
\providecommand{\url}[1]{#1}
\csname url@samestyle\endcsname
\providecommand{\newblock}{\relax}
\providecommand{\bibinfo}[2]{#2}
\providecommand{\BIBentrySTDinterwordspacing}{\spaceskip=0pt\relax}
\providecommand{\BIBentryALTinterwordstretchfactor}{4}
\providecommand{\BIBentryALTinterwordspacing}{\spaceskip=\fontdimen2\font plus
\BIBentryALTinterwordstretchfactor\fontdimen3\font minus \fontdimen4\font\relax}
\providecommand{\BIBforeignlanguage}[2]{{%
\expandafter\ifx\csname l@#1\endcsname\relax
\typeout{** WARNING: IEEEtran.bst: No hyphenation pattern has been}%
\typeout{** loaded for the language `#1'. Using the pattern for}%
\typeout{** the default language instead.}%
\else
\language=\csname l@#1\endcsname
\fi
#2}}
\providecommand{\BIBdecl}{\relax}
\BIBdecl

\bibitem{Fleckenstein2021AgeRelatedMacularDegeneration}
M.~Fleckenstein, T.~D.~L. Keenan, R.~H. Guymer, U.~Chakravarthy, S.~Schmitz‑Valckenberg, C.~C. Klaver, W.~T. Wong, and E.~Y. Chew, ``Age‑related macular degeneration,'' \emph{Nature Reviews Disease Primers}, vol.~7, no.~1, p.~31, 2021.

\bibitem{Antonetti2021MolecularCellularDiabeticRetinopathy}
D.~A. Antonetti, P.~S. Silva, and A.~W. Stitt, ``Current understanding of the molecular and cellular pathology of diabetic retinopathy,'' \emph{Nature Reviews Endocrinology}, vol.~17, no.~4, pp. 195--206, Jan 2021, published online 19 January 2021.

\bibitem{Casson2012DefinitionGlaucoma}
R.~Casson, Glyn Chidlow, J.~Wood, J.~Crowston, and I.~Goldberg, ``Definition of glaucoma: Clinical and experimental concepts,'' \emph{Clinical \& Experimental Ophthalmology}, vol.~40, no.~4, pp. 341--349, Feb 2012.

\bibitem{mandal2021assessing}
S.~Mandal, A.~A. Jammal, and F.~A. Medeiros, ``Assessing glaucoma in retinal fundus photographs using deep feature consistent variational autoencoders,'' \emph{arXiv preprint arXiv:2110.01534}, 2021.

\bibitem{besenczi2016review}
R.~Besenczi, J.~T{\'o}th, and A.~Hajdu, ``A review on automatic analysis techniques for color fundus photographs,'' \emph{Computational and structural biotechnology journal}, vol.~14, pp. 371--384, 2016.

\bibitem{huang2022fully}
S.-Y. Huang, W.-L. Hsu, R.-J. Hsu, and D.-W. Liu, ``Fully convolutional network for the semantic segmentation of medical images: A survey,'' \emph{Diagnostics}, vol.~12, no.~11, p. 2765, 2022.

\bibitem{septiarini2023automatic}
A.~Septiarini, H.~Hamdani, E.~Setyaningsih, E.~Junirianto, and F.~Utaminingrum, ``Automatic method for optic disc segmentation using deep learning on retinal fundus images,'' \emph{Healthcare Informatics Research}, vol.~29, no.~2, pp. 145--151, 2023.

\bibitem{xiao2023transformers}
H.~Xiao, L.~Li, Q.~Liu, X.~Zhu, and Q.~Zhang, ``Transformers in medical image segmentation: A review,'' \emph{Biomedical Signal Processing and Control}, vol.~84, p. 104791, 2023.

\bibitem{ravi2024sam2segmentimages}
\BIBentryALTinterwordspacing
N.~Ravi, V.~Gabeur, Y.-T. Hu, R.~Hu, C.~Ryali, T.~Ma, H.~Khedr, R.~Rädle, C.~Rolland, L.~Gustafson, E.~Mintun, J.~Pan, K.~V. Alwala, N.~Carion, C.-Y. Wu, R.~Girshick, P.~Dollár, and C.~Feichtenhofer, ``Sam 2: Segment anything in images and videos,'' 2024. [Online]. Available: \url{https://arxiv.org/abs/2408.00714}
\BIBentrySTDinterwordspacing

\bibitem{zhang2024unleashingpotentialsam2biomedical}
\BIBentryALTinterwordspacing
Y.~Zhang and Z.~Shen, ``Unleashing the potential of sam2 for biomedical images and videos: A survey,'' 2024. [Online]. Available: \url{https://arxiv.org/abs/2408.12889}
\BIBentrySTDinterwordspacing

\bibitem{zhu2024medicalsam2segment}
\BIBentryALTinterwordspacing
J.~Zhu, A.~Hamdi, Y.~Qi, Y.~Jin, and J.~Wu, ``Medical sam 2: Segment medical images as video via segment anything model 2,'' 2024. [Online]. Available: \url{https://arxiv.org/abs/2408.00874}
\BIBentrySTDinterwordspacing

\bibitem{hu2021lora4}
E.~Hu, Y.~Shen, and H.~Sun, ``Lora: Low-rank adaptation of large language models,'' \emph{arXiv preprint arXiv:2106.09685}, 2021.

\bibitem{chen2024hiera}
C.~Chen, J.~Biffle, X.~Chen, A.~Kirillov, E.~Mintun, X.~Zhu, P.~Dollár, and R.~Girshick, ``Hiera: A hierarchical vision transformer without the bells-and-whistles,'' in \emph{The International Conference on Learning Representations (ICLR)}, 2024.

\bibitem{kirillov2023segment}
A.~Kirillov, E.~Mintun, N.~Ravi, H.~Mao, C.~Rolland, L.~Gustafson, T.~Xiao, S.~Whitehead, A.~C. Berg, W.-Y. Lo \emph{et~al.}, ``Segment anything,'' in \emph{Proceedings of the IEEE/CVF international conference on computer vision}, 2023, pp. 4015--4026.

\bibitem{long2015fully7}
J.~Long, E.~Shelhamer, and T.~Darrell, ``Fully convolutional networks for semantic segmentation,'' in \emph{Proceedings of the IEEE Conference on Computer Vision and Pattern Recognition}, 2015.

\bibitem{milletari2016v}
F.~Milletari, N.~Navab, and S.-A. Ahmadi, ``V-net: Fully convolutional neural networks for volumetric medical image segmentation,'' in \emph{2016 fourth international conference on 3D vision (3DV)}.\hskip 1em plus 0.5em minus 0.4em\relax Ieee, 2016, pp. 565--571.

\bibitem{abraham2019novel}
N.~Abraham and N.~M. Khan, ``A novel focal tversky loss function with improved attention u-net for lesion segmentation,'' in \emph{2019 IEEE 16th international symposium on biomedical imaging (ISBI 2019)}.\hskip 1em plus 0.5em minus 0.4em\relax IEEE, 2019, pp. 683--687.

\bibitem{fraz2012ensemble}
M.~M. Fraz, P.~Remagnino, A.~Hoppe, B.~Uyyanonvara, A.~R. Rudnicka, C.~G. Owen, and S.~A. Barman, ``An ensemble classification-based approach applied to retinal blood vessel segmentation,'' \emph{IEEE Transactions on Biomedical Engineering}, vol.~59, no.~9, pp. 2538--2548, 2012.

\bibitem{staal2004ridge}
J.~Staal, M.~D. Abr{\`a}moff, M.~Niemeijer, M.~A. Viergever, and B.~Van~Ginneken, ``Ridge-based vessel segmentation in color images of the retina,'' \emph{IEEE transactions on medical imaging}, vol.~23, no.~4, pp. 501--509, 2004.

\bibitem{jin2022fives}
K.~Jin, X.~Huang, J.~Zhou, Y.~Li, Y.~Yan, Y.~Sun, Q.~Zhang, Y.~Wang, and J.~Ye, ``Fives: A fundus image dataset for artificial intelligence based vessel segmentation,'' \emph{Scientific data}, vol.~9, no.~1, p. 475, 2022.

\bibitem{budai2013robust}
A.~Budai, R.~Bock, A.~Maier, J.~Hornegger, and G.~Michelson, ``Robust vessel segmentation in fundus images,'' \emph{International journal of biomedical imaging}, vol. 2013, no.~1, p. 154860, 2013.

\bibitem{hoover2000locating}
A.~Hoover, V.~Kouznetsova, and M.~Goldbaum, ``Locating blood vessels in retinal images by piecewise threshold probing of a matched filter response,'' \emph{IEEE Transactions on Medical imaging}, vol.~19, no.~3, pp. 203--210, 2000.

\bibitem{6867807}
J.~Sivaswamy, S.~R. Krishnadas, G.~Datt~Joshi, M.~Jain, and A.~U. Syed~Tabish, ``Drishti-gs: Retinal image dataset for optic nerve head(onh) segmentation,'' in \emph{2014 IEEE 11th International Symposium on Biomedical Imaging (ISBI)}, 2014, pp. 53--56.

\bibitem{bajwa2006g1020}
M.~N. Bajwa, G.~A.~P. Singh, W.~Neumeier, M.~I. Malik, A.~Dengel, and S.~Ahmed, ``G1020: a benchmark retinal fundus image dataset for computer-aided glaucoma detection. arxiv. 2020; 2006.09158 [eess. iv]. published online may 28, 2020, doi: 10.48550,'' \emph{arXiv}, 2006.

\bibitem{huang2023grape}
X.~Huang, X.~Kong, Z.~Shen, J.~Ouyang, Y.~Li, K.~Jin, and J.~Ye, ``Grape: A multi-modal dataset of longitudinal follow-up visual field and fundus images for glaucoma management,'' \emph{Scientific Data}, vol.~10, no.~1, p. 520, 2023.

\bibitem{zhang2010origa}
Z.~Zhang, F.~S. Yin, J.~Liu, W.~K. Wong, N.~M. Tan, B.~H. Lee, J.~Cheng, and T.~Y. Wong, ``Origa-light: An online retinal fundus image database for glaucoma analysis and research,'' in \emph{2010 Annual international conference of the IEEE engineering in medicine and biology}.\hskip 1em plus 0.5em minus 0.4em\relax IEEE, 2010, pp. 3065--3068.

\bibitem{kovalyk2022papila}
O.~Kovalyk, J.~Morales-S{\'a}nchez, R.~Verd{\'u}-Monedero, I.~Sell{\'e}s-Navarro, A.~Palaz{\'o}n-Cabanes, and J.-L. Sancho-G{\'o}mez, ``Papila: Dataset with fundus images and clinical data of both eyes of the same patient for glaucoma assessment,'' \emph{Scientific Data}, vol.~9, no.~1, p. 291, 2022.

\bibitem{fang2202refuge2}
H.~Fang, F.~Li, J.~Wu, H.~Fu, X.~Sun, J.~Son, S.~Yu, M.~Zhang, C.~Yuan, and C.~Bian, ``Refuge2 challenge: A treasure trove for multi-dimension analysis and evaluation in glaucoma screening.'' \emph{arXiv preprint arXiv:2202.08994}, 2022.

\bibitem{lu2023empirical}
Y.~Lu, C.~Li, H.~Liu, J.~Yang, J.~Gao, and Y.~Shen, ``An empirical study of scaling instruct-tuned large multimodal models,'' \emph{arXiv preprint arXiv:2309.09958}, 2023.

\bibitem{shen2022self}
X.~Shen, J.~Xu, H.~Jia, P.~Fan, F.~Dong, B.~Yu, and S.~Ren, ``Self-attentional microvessel segmentation via squeeze-excitation transformer unet,'' \emph{Computerized Medical Imaging and Graphics}, vol.~97, p. 102055, 2022.

\bibitem{cao2023mfa}
J.~Cao, J.~Chen, Y.~Gu, and J.~Liu, ``Mfa-unet: A vessel segmentation method based on multi-scale feature fusion and attention module,'' \emph{Frontiers in Neuroscience}, vol.~17, p. 1249331, 2023.

\bibitem{seo2025full}
S.~Seo, H.~Yoon, S.~Kim, and J.~Lee, ``Full-scale representation guided network for retinal vessel segmentation,'' \emph{arXiv preprint arXiv:2501.18921}, 2025.

\bibitem{yii2023data}
F.~Yii, T.~MacGillivray, and M.~O. Bernabeu, ``Data efficiency of segment anything model for optic disc and cup segmentation,'' in \emph{International conference on medical image computing and computer-assisted intervention}.\hskip 1em plus 0.5em minus 0.4em\relax Springer, 2023, pp. 336--346.

\bibitem{10551123}
C.~Meas, W.~Guo, and M.~H. Miah, ``Multi-scale attention u-net for optic disc and optic cup segmentation in retinal fundus images,'' in \emph{2024 2nd International Conference on Advancement in Computation \& Computer Technologies (InCACCT)}, 2024, pp. 760--765.

\bibitem{chen2024optic}
Y.~Chen, Y.~Bai, and Y.~Zhang, ``Optic disc and cup segmentation for glaucoma detection using attention u-net incorporating residual mechanism,'' \emph{PeerJ Computer Science}, vol.~10, p. e1941, 2024.

\bibitem{tadisetty2023identifying}
S.~Tadisetty, R.~Chodavarapu, R.~Jin, R.~J. Clements, and M.~Yu, ``Identifying the edges of the optic cup and the optic disc in glaucoma patients by segmentation,'' \emph{Sensors}, vol.~23, no.~10, p. 4668, 2023.

\bibitem{bajwa2020g1020}
M.~N. Bajwa, G.~A.~P. Singh, W.~Neumeier, M.~I. Malik, A.~Dengel, and S.~Ahmed, ``G1020: A benchmark retinal fundus image dataset for computer-aided glaucoma detection,'' in \emph{2020 International Joint Conference on Neural Networks (IJCNN)}.\hskip 1em plus 0.5em minus 0.4em\relax IEEE, 2020, pp. 1--7.

\bibitem{sam_zhu2024medical}
J.~Zhu, A.~Hamdi, Y.~Qi, Y.~Jin, and J.~Wu, ``Medical sam 2: Segment medical images as video via segment anything model 2,'' \emph{arXiv preprint arXiv:2408.00874}, 2024.

\end{thebibliography}

\end{document}